\documentclass[11pt]{article} 
\usepackage{rldmsubmit,palatino}
\usepackage{graphicx}

\usepackage{times}
\usepackage{helvet}
\usepackage{courier}

\usepackage{amssymb}
\usepackage{algorithm}
\usepackage{algorithmic}

\usepackage{caption}
\usepackage{subcaption}

\newcommand{\UR}{UR} 
\newcommand{\GE}{GE} 
\newcommand{\BE}{BE} 

\title{Actively Learning to Attract Followers on Twitter}

\author{
Nir Levine \\
Electrical Engineering\\
The Technion\\
Haifa, Israel \\
\texttt{levinir@campus.technion.ac.il} \\
\And
Timothy A. Mann \\
Electrical Engineering\\
The Technion \\
Haifa, Israel \\
\texttt{mann@ee.technion.ac.il}
\AND
 \\
 \\
 \\
\texttt{} \\
\And
Shie Mannor \\
Electrical Engineering \\
The Technion \\
Haifa, Israel \\
\texttt{shie@ee.technion.ac.il} \\
\And
 \\
\\
 \\
\texttt{} \\
\\
}

\begin{document}

\maketitle

\begin{abstract}
Twitter, a popular social network, presents great opportunities for on-line machine learning research. However, previous research has focused almost entirely on learning from passively collected data. We study the problem of learning to acquire followers through normative user behavior, as opposed to the mass following policies applied by many bots. We formalize the problem as a contextual bandit problem, in which we consider retweeting content to be the action chosen and each tweet (content) is accompanied by context. We design reward signals based on the change in followers. The result of our month long experiment with 60 agents suggests that (1) aggregating experience across agents can adversely impact prediction accuracy and (2) the Twitter community's response to different actions is non-stationary. Our findings suggest that actively learning on-line can provide deeper insights about how to attract followers than machine learning over passively collected data alone.
\end{abstract}


\acknowledgements{The research leading to these results has received funding from the European Research Council under the European Union’s Seventh Framework Programme (FP/2007-2013) / ERC Grant Agreement n.306638.
}

\startmain 

\section{Introduction}

Twitter is an on-line social network that allows people to post short messages (140 characters maximum), called status updates or tweets. It also allows users to read status updates posted by other users. Two additional actions are: (1) re-posting another user's status update, this action is called retweeting, and (2) favoriting a tweet, i.e. the tweet is marked as a favorite of the user. As one of the largest social networks, Twitter provides a great opportunity for evaluating machine learning algorithms on real-world data and evaluating them on-line. We target the problem of attracting followers in a community on Twitter and argue that actively learning can provide deeper insights than learning over passively collected data alone.

Without performing active experiments it is difficult to determine whether factors turned up in the analysis are only correlated, but not causally related, with attracting followers. Moreover these factors may change over time or vary depending on the history of an individual user. Therefore running experiments where algorithms control an account (rather than simply observing its behavior) can provide useful insights into the development of on-line relationships.

One difficulty of creating agents with the goal of acquiring followers is that naive exploitative strategies, such as mass following, are quite successful. This kind of aggressive following policy is easily labeled as bot behavior and is treated accordingly by the Twitter community. Our objective is to learn strategies that attract followers but avoid violating behavioral norms. By doing so, the agents receive followers based on providing a valuable service rather than exploiting other users.

Twitter has been the subject of intense research. Most of the work so far applied machine learning on passively collected data as opposed to learning from data collected on-line. However, collecting data on-line allows the agent to respond to the social network in real-time and choose actions most suitable to that time. This allows us to learn what causes users to follow (not just what is correlated). Although the problem of acquiring followers seems to be a popular subject on the Internet, we are unaware of any academic research that has examined an on-line approach for learning to attract followers.

The main contributions of this work are:

(1) We formalized the problem for learning to attract followers as a contextual bandit problem \cite{Li2010}. Our formulation encourages normative (rather than exploitative) behaviors, because the action space focuses on retweeting content (not following users).

(2) We executed a month long on-line experiment with 60 agents. Each agent interacted directly with the Twitter API and controlled a live Twitter account.

(3) We provide evidence for advantages of active learning over learning from passively collected data: the reward signal, based on the change in followers is non-stationary. Analyzing a data set collected in the past may result in poor performance, and more data is not always better. Although we had limited data for each individual agent, we found that aggregating data from multiple agents resulted in less accurate prediction. Therefore, learning should be applied to individual agents.

\section{The Setting}

One of the main contributions of this work is the development of a well-defined problem for actively learning in a social network. To our knowledge, no previous work has proposed a well-defined, active problem in social networks.

We allow a learning system to control a single user account on the Twitter social network. We will refer to this learning system as an agent. Our first intuition was to identify a multi-armed bandit problem \cite{Lai1985}, where the agent plays in a series of rounds. Let $A$ be a set of $K\geq 2$ actions. At each round, the agent selects an action $a \in A$ associated with an initially unknown probability distribution over rewards. Once an action is selected, a reward is sampled from its corresponding distribution. The goal of the agent is to maximize its expected reward. 

To formalize a problem as a multi-armed bandit, we needed to answer a few questions: (1) What are the rewards? (2) What are the actions available to the agent? (3) What is a round?

\subsubsection{Rewards} Our objective is to design an agent that learns how to acquire followers. However, gaining followers is a rare event. A reward based entirely on the change in followers is difficult to learn from because the signal is sparse. Thus, we constructed reward signals from both changes in number of followers and other related events as explained in section \ref{sec:exp}.

\subsubsection{Actions} The number of messages that can be expressed in 140 characters is too large to address directly. Additionally, users can perform other actions like following other users, retweeting existing status updates, and favoriting tweets, thereby significantly increasing the number of possible actions. However, a traditional multi-armed bandit problem assumes that the number of actions $K$ is reasonably small.

In our experiment we used other users' content as actions rather than trying to invent new content. Furthermore, we restricted the set of tweets considered by the agent for retweeting to recent tweets containing the word ``baseball''. We chose the domain ``baseball'' because it has a large community but still represents a specialized domain where agents can learn to retweet valuable content.

Since in our setting it is very unlikely to sample more than once a specific action, we modeled the problem as a contextual bandit problem \cite{Li2010}, which is a generalization of the multi-armed bandit problem. In each trial $t$, the algorithm observes a set $\mathcal{A}_t$ of possible actions sampled from a distribution $\rho$ with support $\mathcal{A}$, where $\mathcal{A}$ is the collection of all tweets. Each action $a \in \mathcal{A}_t$ corresponds to a status update that can be retweeted and is described by a feature vector $x_{t,a}$. This feature vector $x_{t,a}$ describes properties of the action and the state of the environment. After choosing an action, the algorithm receives a reward $r_{t,a_t}$ from which it learns and improves to the next trial. The objective of the agent is to learn a decision rule $\mu$ that maximizes
\begin{equation}
r = \mathbb{E}_{\mathcal{A}_t \sim \rho} \left[ r_{t,a_t} \mid \mu \right] \enspace ,
\end{equation}
where $r$ is the expected reward. In our experiment we use decision rules of the form
\begin{equation}
\mu (\mathcal{A}_t ) = \arg \max_{a \in \mathcal{A}_t } f(x_{t,a})
\end{equation}
where $f$ is a learned function that predicts the expected reward.

Due to the high-dimension of $X=\{x_{t,a}\}$, the feature space, we estimate rewards using function approximation/regression. The result of function approximation is a function $f : X \rightarrow \mathbb{R}$ that predicts the reward received for selecting the action $a \in A_t$ corresponding to $x_{t,a}$.

\subsubsection{Rounds} A round could potentially be any constant length of time. However, enforcing normative behavior along with providing enough time between actions to allow other users to respond, led us to choose a time period of one hour between rounds.

A typical round in our setting is as following. First, the agent requests and receives a collection of recent tweets from the Twitter API about a specialized domain (e.g., baseball). Next, the agent examines the tweets it received from the Twitter API and retweets one of them. Finally, the agent sleeps for an hour and requests information about what changed. The agent then uses this information to calculate a reward and updates its decision rule. This finishes the round and a new round begins immediately.

\subsubsection{The Exploration-Exploitation Dilemma} Exploration, trying actions with uncertain reward, is a critical issue in multi-armed bandit problems. We want the agent to choose an action that will have a high probability of attracting followers (exploitation), but the agent also needs to try various actions to learn what attracts followers (exploration). For simplicity, we use a simple the popular $\epsilon$-greedy exploration strategy, which selects a random action with probability $\epsilon$ and the action with the highest predicted reward with probability $(1-\epsilon)$.

\section{Experiment}
\label{sec:exp}

We designed and executed an experiment to determine whether simple machine learning algorithms could learn to attract more followers on Twitter than a random baseline. We created 60 Twitter accounts, each agent controlled one account. Every hour $t \geq 0$, each agent requested a collection of tweets $\mathcal{A}_t$ from Twitter. The agents selected a tweet $a_t \in \mathcal{A}_t$ to be retweeted (i.e., a status update) based on a list of features that were extracted from the tweets, $x_{t,a_t}$. One hour later a reward signal for $a_t$ was computed by the agent. In line with our objective of maintaining normative behavior, agents only followed the user that posted $a_t$ (before the agent retweeted it) with probability $P(follow)=0.5$. The entire experiment was performed by using Twitter API.\footnote{\texttt{https://dev.twitter.com/docs/api/1.1}}

The reward signal used during the experiment was 
\begin{equation} \label{eqn:reward}
r_{t,a_t} = \alpha_0 \Delta_{a,t} + \alpha_1 \Delta_{u,t} + \alpha_2 f_t + \alpha_3 w_t \enspace ,
\end{equation}
where $\Delta_{a,t}$ is the change in the number of agent's followers, $\Delta_{u,t}$ is the change in the number of followers for the tweet's original poster, $f_t$ is the number of favorites the tweet received, and $w_t$ is the number of retweets made to this tweet. The coefficients were $\alpha_0 = 100, \alpha_1 = 10, \alpha_2 = 10,$ and $\alpha_3 = 1$, aligned with our objective we gave significantly higher weight to change in number of agent's followers.

Our experiment focused on tweets containing the string ``baseball'', because there is a constant flow of status updates and the topic is specialized enough so that an agent might learn useful knowledge about the domain. Status updates with offensive language were filtered before the agent made a selection. We divided the agents in three groups of 20: (1) uniform random (\UR), (2) a gradient-based estimator (\GE), and (3) a batch-based estimator (\BE). 

\GE\ and \BE\ estimated the reward signal for each status update from a collection of 87 features. 
We extracted features from the tweet and the user that posted it. We selected features based on prior work, such as \cite{Zarrella2009,Suh2010}, along with others.

To encourage exploration, \GE\ and \BE\ selected a status update according to the uniform random rule with probability $\epsilon = 0.05$. 

\subsection{Uniform Random (\UR) Agents}
The baseline \UR\ algorithm selected retweets according to a uniform random distribution over the set of status updates. Thus, \UR\ does not do any learning. 

\subsection{Gradient Estimator (\GE) Agents}
The \GE\ algorithm incrementally updated a linear function approximator. \GE\ applied gradient descent to minimize Mean Squared Error (MSE) and used a constant learning rate ($\eta = 0.1$). To better utilize the data, we introduced an adviser \cite{papudesi2003} to the \GE\ algorithm. The adviser observes tweets not retweeted by the agent, and one hour later, at the same time the agent observes its reward, the adviser computes a modified reward signal
\begin{equation} \label{eqn:reward_adv}
r'_{t,a_t} = \beta_0 \Delta_{a,u_t} + \beta_1 f_{a,t} + \beta_2 w_{a,t} \enspace ,
\end{equation}
for the other tweets where $\Delta_{a,u_t}$ is the change in the number of followers for the tweet's original poster, $f_{a,t}$ is number of favorites the tweet received, $w_{a,t}$ is the number of retweets made to this tweet, and coefficients $\beta_0 = 10, \beta_1 = 10, \beta_2 = 1$. Then, the adviser generates a hypothesis, using a batch training approach. The hypothesis generated by the adviser is weighted with the hypothesis generated by the agent. To reduce noise in the reward signal, \GE\ only takes a learning step each 8 hours, averaging over all the hypotheses, to generate a new hypothesis as suggested by \cite{dekel2005}.


\subsection{Batch-based Estimator (\BE) Agents}
The \BE\ algorithm used Ordinary Least Squares (OLS) to train a linear function approximator after each round on all instances where the algorithm received the reward signal (\ref{eqn:reward}). All samples were weighted the same and the algorithm minimized MSE over the training samples. The \BE\ algorithm used only samples it had selected to retweet (i.e., it did not use an advisor like the \GE\ agents).

\section{Experimental Results}

We ran the experiment for about one month (May 9 -- June 11, 2014) generating more than 600 status-updates for each account. At the end of the experiment, the average number of followers for the different groups was: 41.1 for \UR, \BE\ finished with 41.6, and \GE\ had 44.95. Thus, \GE\ acquired about 10\% more followers on average than \UR. A one sided T-test on the \UR\ and \GE\ groups shows that the difference is statistically significant with $p=0.0291$ (i.e., 97.09\% confidence that the two groups are generated by distributions with different means). However, the difference between \UR\ and \BE\ groups was not significant (see the next section for details). Two weeks before the end of the experiment, the average number of followers was 23.6, 23.6, and 23.8 for \BE, \UR, and \GE\ respectively. However, over the last two weeks the difference between \UR\ and \GE\ grew 11 fold. This demonstrates that machine learning can attract more followers than a random strategy. It also raises the question: Why did \GE\ outperform \UR, while \BE\ did not? In the next section, we analyze our experimental data to gain insight on attracting followers.

\section{Importance of Active Learning}

For comparison, throughout the analysis we normalized the rewards to have zero mean and a standard deviation of one. We noticed the poor performance of the \BE\ algorithm. When we tested the performance offline, we found that Ordinary Least Squares (OLS) resulted in divergence. Therefore, while analyzing the data we used different function approximation models, Ridge Regression, LASSO, Elastic Net, and Support Vector Regression (SVR).\footnote{The implementations used in our analysis are available at \texttt{http://scikit-learn.org}.} These methods apply types of regularization, thus gave better results (in off-line training).

In this section we show the importance of actively learning from data as opposed to learning from passively collected data. We identify two main problems in learning from passively collected: (1) the relationship between the features and reward function appears to be non-stationary, and (2) generalizing between agents tends to degrade prediction accuracy.

\begin{figure}
\centering
\begin{minipage}{.4\textwidth}
  \centering
  \includegraphics[width=0.9\textwidth]{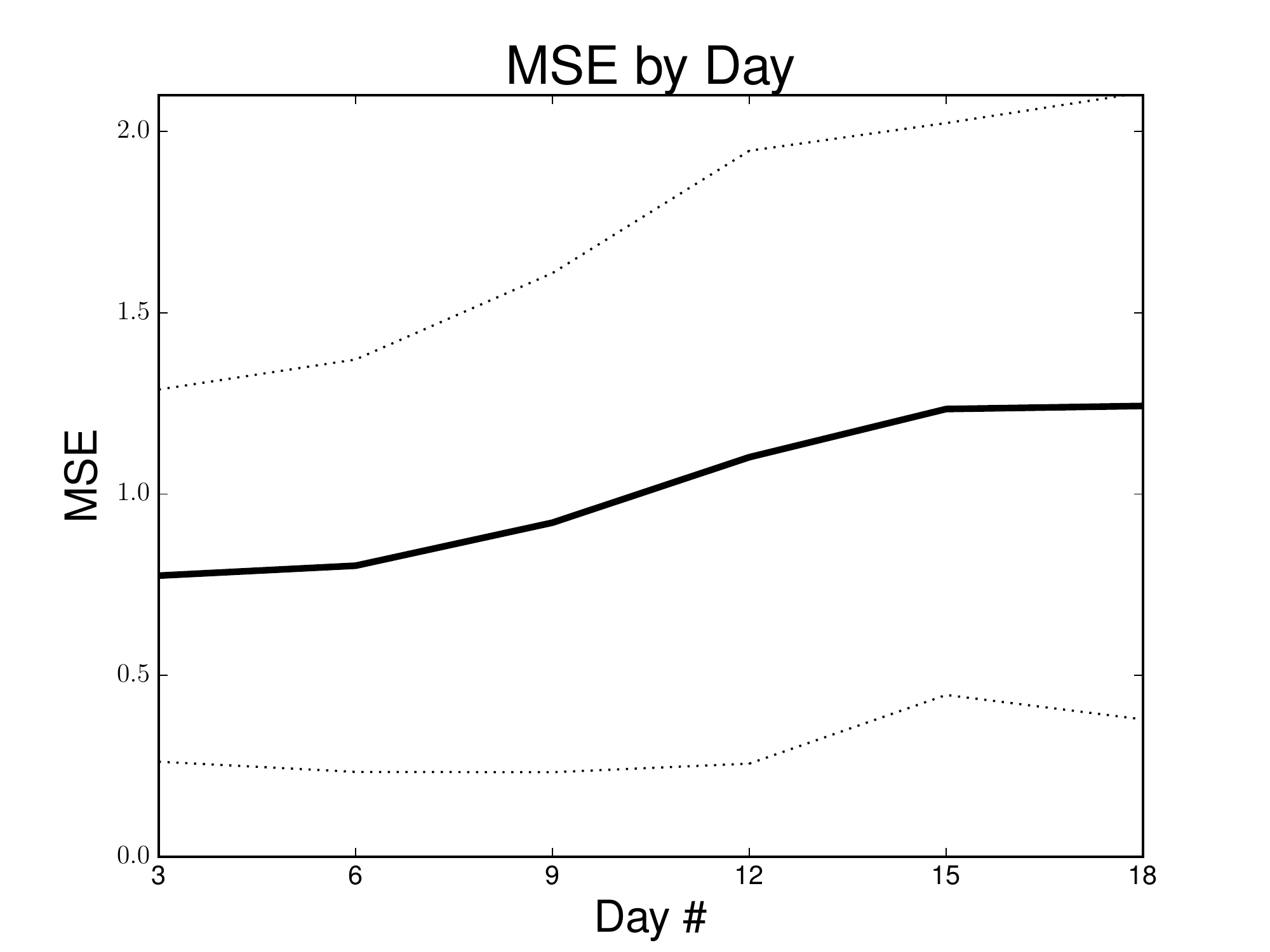}
  \captionof{figure}{MSE of an SVR algorithm trained on the first 100 samples and tested on the remainder of the data. The MSE is plotted as a function of time (in days) after the 100 samples were collected. The error increases over time indicating that the reward signal we are trying to predict is nonstationary.}
  \label{fig:nonstationary}
\end{minipage}
\hspace{.05\linewidth}
\begin{minipage}{.4\textwidth}
  \centering
  \includegraphics[width=0.85\textwidth]{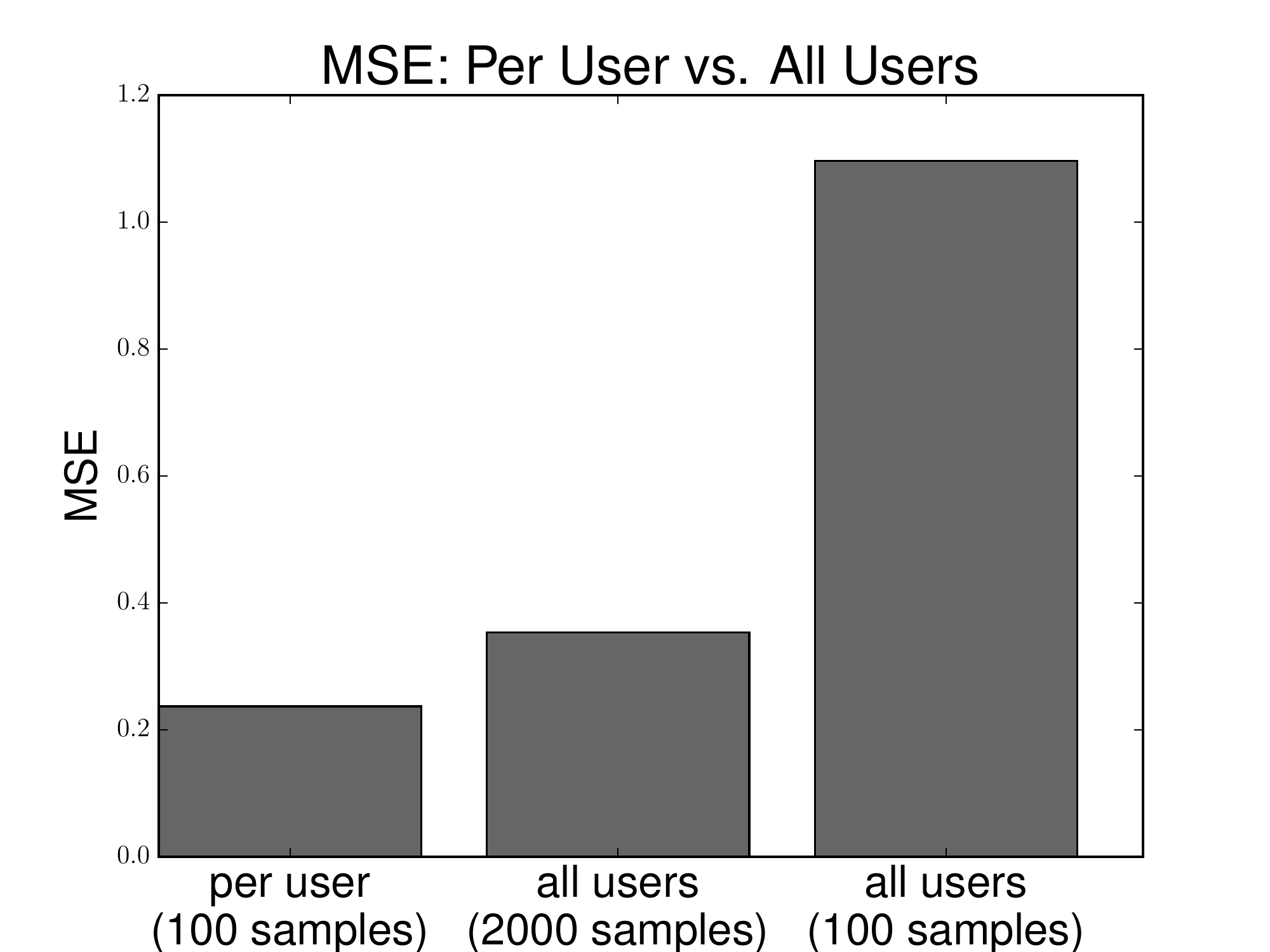}
  \captionof{figure}{Comparison of MSE per user with 100 samples, all users with 2000 samples, and all users with 100 samples (75\% training, 25\% test). Combining data between users negatively impacts prediction accuracy.}
  \label{fig:per_user_vs_all_users}
\end{minipage}
\end{figure}

\subsection{Nonstationarity}

By examining the data, we found considerable evidence that the reward signal is a non-stationary function of our features. Thus training only on passively collected data is probably not satisfactory, because the best actions for acquiring followers seem to be time sensitive.

For each \BE\ agent data (sorted chronologically) we trained and evaluated SVR model (achieved the smallest MSE). We divided the samples into chunks containing 100 sequential instances. The size was selected by experimentation yielding the smallest MSE. Each chunk was split into 75\% training data and 25\% testing data. Finally, we took the median MSE over all of the agents. The median MSE was 0.24, an improvement of 15\% over when trained with all the data together. Thus, training with all data resulted in more error.

Next we looked at predicting the number of followers rather than the reward signal (\ref{eqn:reward}) used in our experiments. For each agent, we trained SVR on the first 100 samples (sorted chronologically) and then used the remaining samples as test data. Figure \ref{fig:nonstationary} shows that the MSE increases as the experiment progresses. This is consistent with our hypothesis that the best strategy for attracting followers is changing over time. 

These findings suggests that the reward signal is non-stationary, therefore learning from on-line data may result in more accurate predictions, as opposed to learning from passively collected data.

\subsection{Generalizing Across Users}

We examined the evolution of the weights learned by the \GE\ agents. Specifically, we examined the median values and standard deviations  for the weights between all the agents. The median values did not converge to a single point on weights space. On the contrary, the standard deviations increased over time, meaning the agents were learning different hypotheses.

Figure \ref{fig:per_user_vs_all_users} shows the increase in error for \BE\ agents when generalizing between agents compared to training on a single agent's data. We show the MSE for an SVR algorithm in three different cases, (1) per user with 100 samples, (2) combined data with 2000 samples, (3) combined data with 100 samples (using a moving window over the data). The MSE increases when generalizing over data from multiple users. When we used the same sample size (100) the MSE increased dramatically (around 4 fold). Even when we used a 20 times bigger training sets for the combined users' data (corresponding to 100 samples from each agent), the MSE still increased compared to the per user setup.

This demonstrates the importance of an agent learning from its own history. Simply generalizing over multiple agents' history actually degrades performance even with significantly more training data.

\bibliographystyle{plain}
\bibliography{twitter}

\end{document}